\documentclass{article} 
\usepackage{iclr2026_conference,times}

\usepackage{amsmath,amsfonts,bm}

\def\eqref#1{equation~\ref{#1}}

\def\1{\bm{1}}

\DeclareMathAlphabet{\mathsfit}{\encodingdefault}{\sfdefault}{m}{sl}
\SetMathAlphabet{\mathsfit}{bold}{\encodingdefault}{\sfdefault}{bx}{n}

\usepackage{hyperref}
\usepackage{url}
\usepackage{booktabs}
\usepackage{graphicx}
\usepackage{xcolor}
\usepackage{float}

\title{Characterizing Brazilian Atlantic Forest Restoration Outcomes with \\Geospatial AlphaEarth Embeddings}

\author{Alice Heiman \\
Department of Computer Science\\
Stanford University\\
\texttt{\{aheiman\}@stanford.edu} \\
}

\iclrfinalcopy 
\begin{document}

\maketitle

\begin{abstract}
The Atlantic Forest in Brazil is a critical biodiversity hotspot, yet less than 12–15\% of its original cover remains. Although monitoring forest restoration on a large scale is essential, traditional methods are limited by the impracticality of on-the-ground reporting on such a scale and by the saturation of remote-sensing indices such as NDVI. Furthermore, reforestation is a gradual process as opposed to the rapid spectral changes caused by deforestation. In this study, we examine 1,729 restoration sites in São Paulo, using satellite embeddings from the AlphaEarth Foundation's model to evaluate their effectiveness in characterising early restoration success. We introduce the concept of a ``Reference Trajectory Embedding'', defining a metric of restoration success based on cosine similarity to reference sites of mature secondary forest. We observe distinct clusters in embedding space according to different land use and land cover (LULC) types, and we can identify sites with clear change vectors. However, the signal can be noisy, and embeddings may require further fine-tuning to capture and predict site metadata beyond LULC.
\end{abstract}

\section{Introduction}

Brazil, short for 'Terra do Brasil' (Eng. 'Land of Brazilwood'), is home to some of the world's most important and biodiverse forests. However, the tree that gave Brazil its name is not found in the Amazon Rainforest. The endangered paubrasila tree can only be found in the Atlantic Forest. This forest provides ecosystem services to over 150 million Brazilians, contains over 20,000 plant species and supplies 62\% of Brazil's hydroelectric power \citep{delima2020erosion, wwf2025atlantic}.  However, less than 12–15\% of the original forest area remains, which is why it has been designated a World Restoration Flagship Project by the United Nations (UN) \citep{undecade2025atlantic}. Initiatives such as the Trinational Atlantic Forest Pact have attracted over 300 organisations with the aim of restoring the forest. Yet, planting alone is insufficient. Effective restoration requires rigorous monitoring to understand what drives ecological success, but traditional vegetation monitoring metrics, such as NDVI, often saturate in dense tropical regions and lack universal thresholds for success \citep{gao2020remote}. Furthermore, remote sensing methods have traditionally focused on deforestation due to the clearer change in the spectral signal, rather than the gradual changes that occur in recovering forests over decades.

Earth observation foundation models (FMs), such as AlphaEarth Foundations, produce learned embeddings that combine multiple data types. These high-dimensional vectors may encode more ecological information than simple indices. In this paper, we propose using geospatial embeddings to characterise reforestation sites. Specifically, we explore whether geospatial embeddings can be used to: (1) Track the early success of restoration projects in tropical settings. (2) Cluster restoration sites in ecologically meaningful ways based on embeddings; and (3) Infer restoration metadata better than features like NDVI, topology, and climate alone. Our contributions include a novel ``Reference Trajectory Embedding'' success metric that uses cosine similarity to mature secondary forest reference sites, as well as a systematic evaluation of 1,729 reforestation polygons in the Atlantic Forest and their restoration trajectories in embedding space.

\section{Data}

\subsection{Study Sites}

We select a subset of 43,139 restoration polygons from the Observatorio da Restauração e Reflorestamento (ORR). Specifically, we filter for Atlantic Forest sites in the state of São Paulo ($n=14,873$) with a restoration start year between 2017 and 2024 ($n=14,526$). We only use sites with an area of at least 1 hectare after applying a -10 m buffer to avoid edge effects ($n=1,729$). We use São Paulo as a case study because it contains the most complete metadata. We limit the analysis to the period 2017–2024, as this was the availability of the pre-computed AlphaEarth embeddings.

\begin{figure*}[t]
\centering
\includegraphics[width=\textwidth]{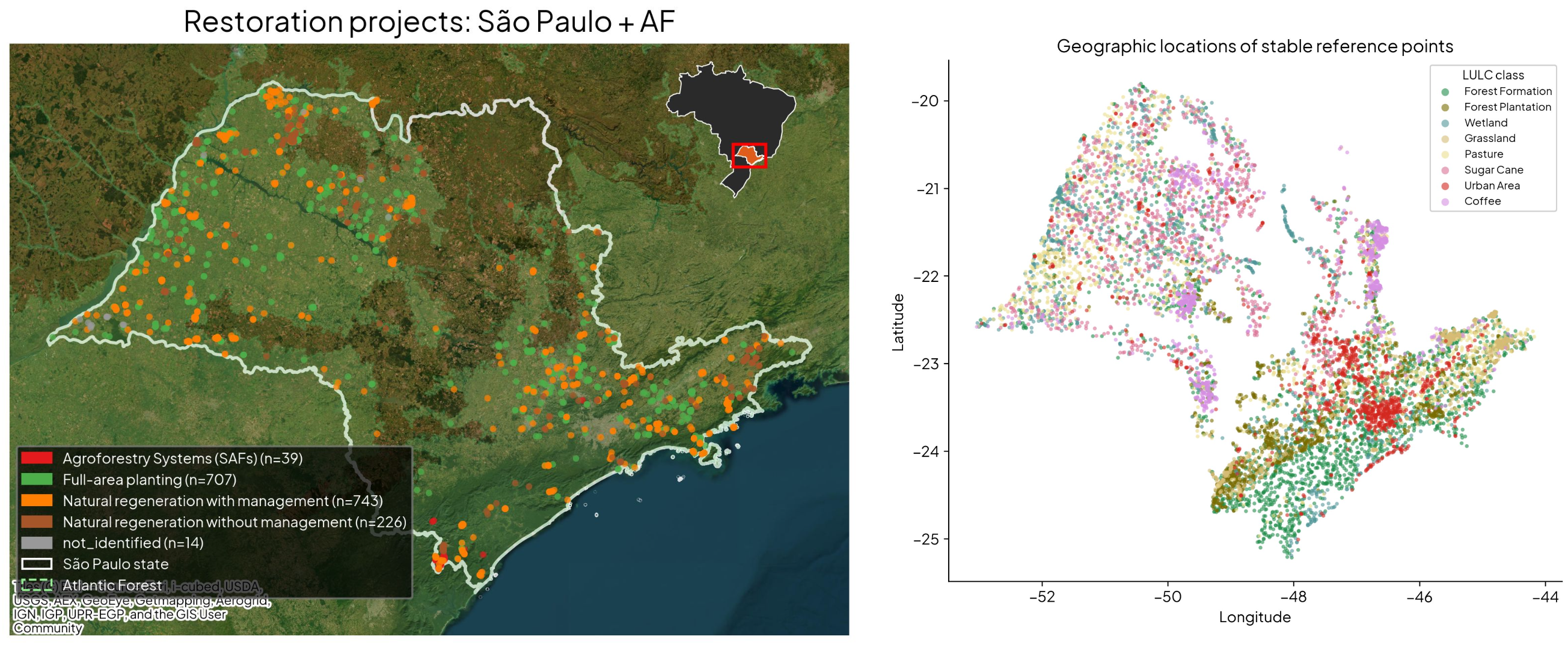}
\caption{Study area in the Atlantic Forest biome, Brazil, with restoration site polygons and reference LULC points marked.}
\label{fig:study-area}
\end{figure*}

\subsection{Reference Point Sampling}

We use the MapBiomas Brazil Collection 10 Coverage V1 and Deforestation Secondary Vegetation V2 layers in Google Earth Engine to sample two categories of reference points. \textit{Stable} points share the same land use and land cover (LULC) class for 10 or more consecutive years through 2024; \textit{changing} points transition from one LULC class (2017-2020) to another (2021-2024). We sample $n{=}1{,}000$ points per class for nine stable classes: Primary Forest, Secondary Forest, Forest Plantation, Wetland, Sugar Cane, Coffee, Grassland, Pasture, and  Urban. We sample $n{=}200$ points per transition for six changing classes: Forest Formation $\rightarrow$ Urban, Forest Formation $\rightarrow$ Sugar Cane, Forest Formation $\rightarrow$ Pasture, Pasture $\rightarrow$ Forest Formation, and Coffee $\rightarrow$ Forest Formation.

\subsection{Data Extraction}

We extract the following data at each reference point or as a restoration polygon mean: \textit{AlphaEarth Embeddings} (64 dimensions per site per year); \textit{Vegetation Indices} (monthly and annual Sentinel-2-derived NDVI and EVI); \textit{Environmental Covariates} (annual precipitation, min/max temperature, elevation, slope, aspect, evapotranspiration, forest cover within 2~km, and road density within 5~km); and \textit{Project Metadata} (restoration area, start year, and strategy from the ORR dataset). See Appendix \ref{sec:appendix_a} for a detailed breakdown of data sources used.

\section{Methods and Results}

\subsection{Embedding Space Visualization}

\begin{figure*}[t]
\centering
\includegraphics[width=\textwidth]{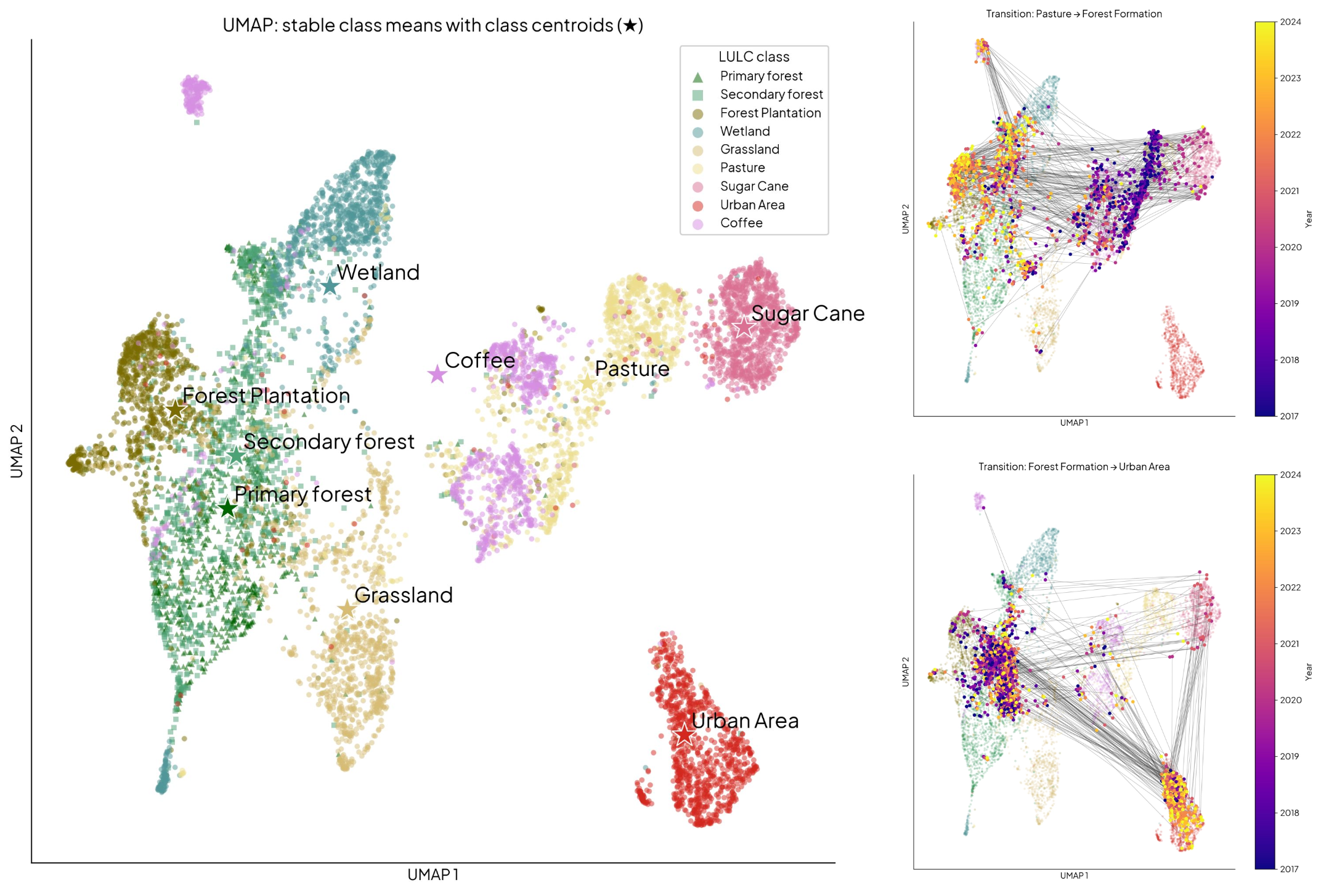}
\caption{AlphaEarth embeddings of reference stable and changing points projected into 2D using UMAP.}
\label{fig:umap_trajectories}
\end{figure*}

To verify the semantics of the AlphaEarth embedding space, we fit a UMAP  on all stable LULC reference point embeddings from 2024 and visualize them  in 2D space. As shown in Figure~\ref{fig:umap_trajectories}, the LULC classes form distinct clusters with interpretable structure: the Urban class is the most separated, while more heterogeneous classes such as 
Forest exhibit greater intra-class spread. Notably, we were able to reveal some misclassifications in the original MapBiomas LULC maps, where points labeled as one class cluster closer to a different class in embedding space (see Appendix \ref{sec:appendix_e}). We further visualize the year-on-year embedding trajectories of known changing reference points to qualitatively verify that transitions between LULC classes are reflected in the embedding space.

\subsection{Restoration Trajectories}

We compute a global reference vector $\bar{\mathbf{r}}$ as the mean embedding of all sampled stable secondary forest pixels. The global similarity of polygon $i$ at year $t$ is
\begin{equation}
  s^{\text{global}}_{i,t}
  = \frac{\mathbf{e}_{i,t} \cdot \bar{\mathbf{r}}}
         {\|\mathbf{e}_{i,t}\|\,\|\bar{\mathbf{r}}\|},
\end{equation}
where $\mathbf{e}_{i,t} \in \mathbb{R}^{64}$ is the polygon's embedding at
year~$t$. To control for regional variation in forest composition, we also compute a \emph{local} reference for each polygon. To do so, we identify the embedding of the closest reference secondary forest point $\bar{\mathbf{r}}_i$ to each restoration polygon. The local similarity $s^{\text{local}}_{i,t}$ is then the cosine similarity between $\mathbf{e}_{i,t}$ and $\bar{\mathbf{r}}_i$. Each polygon thus has a time series $\{s_{i,t}\}_{t=0}^{T_i}$, where $t=0$ corresponds to the restoration start year. To control for different start years, plots show the $\Delta t$, i.e. the years since restoration start year. 

\begin{figure*}[t]
\centering
\includegraphics[width=\textwidth]{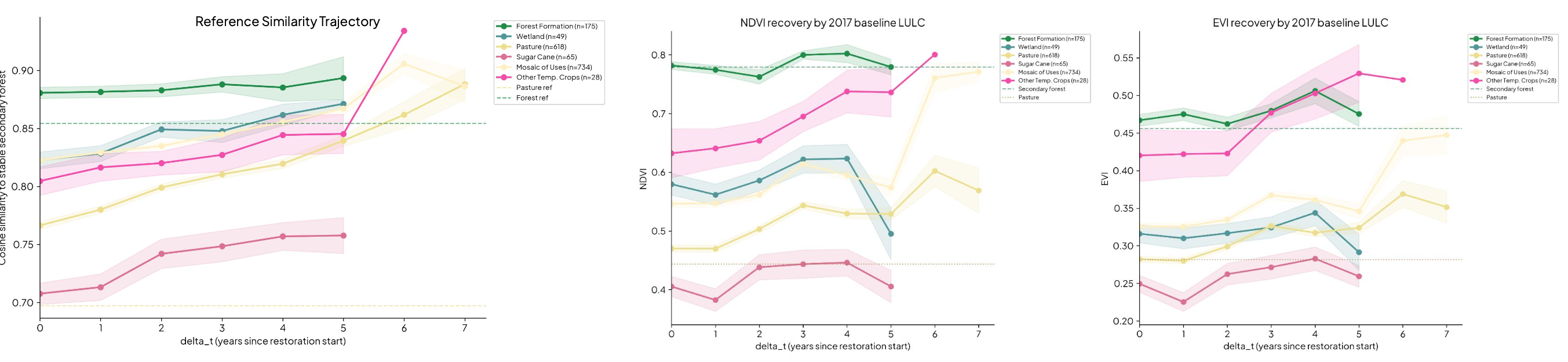}
\caption{Similarity, NDVI, and EVI trajectories stratified by starting MapBiomas LULC class.}
\label{fig:trajectories_lulc}
\end{figure*}

Thus, the similarity scores $s_{i,t}$ can be interpreted as absolute restoration progress scores and visualized as year-on-year trajectories. The difference, meanwhile, can be interpreted as a relative improvement score from the start of the restoration process: $s_{i,T_i} - s_{i,0}$. We use stable-class means (primary forest and pasture) to provide upper and lower similarity baseline references to contextualise the trajectories. We also compare the trajectories to the NDVI and EVI spectral indices (see Figure \ref{fig:trajectories_lulc}).

\section{Prediction Tasks}

\begin{figure*}[t]
\centering
\includegraphics[width=0.9\textwidth]{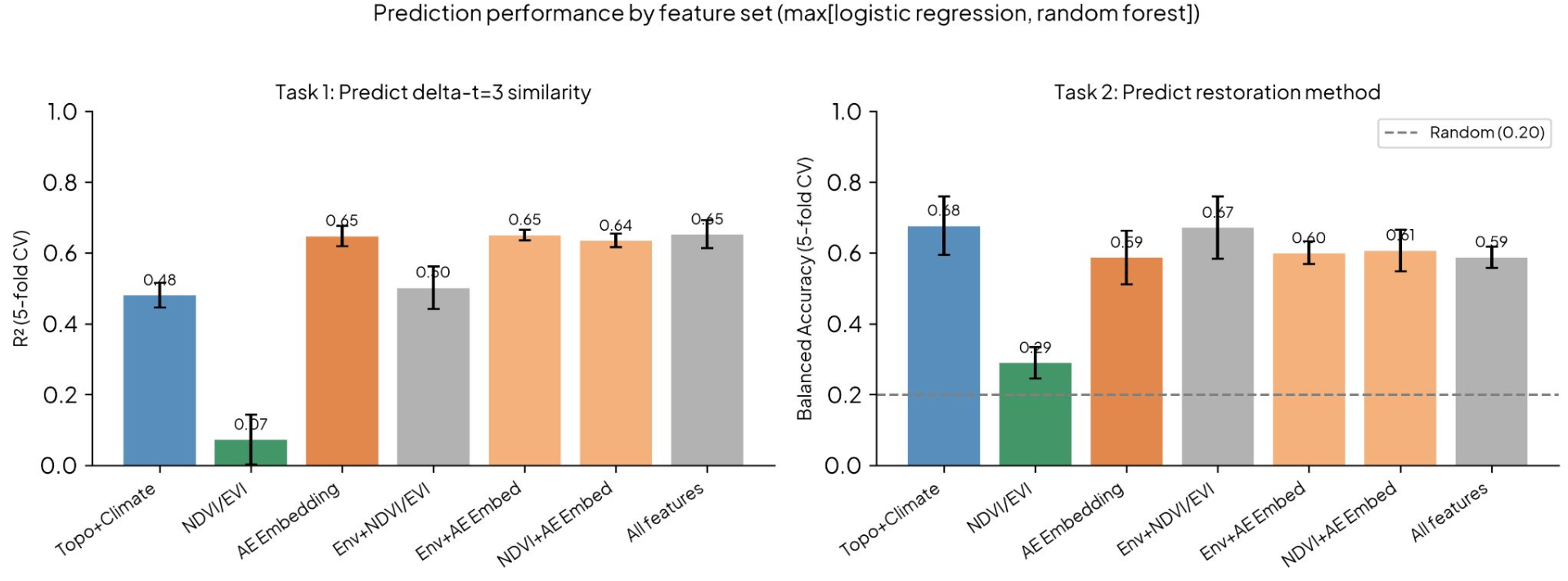}
\caption{Results from predicting future reference similarity and polygon metadata.}
\label{fig:prediction_tasks}
\end{figure*}

We pose two tasks for predicting restoration outcomes: (1) predict similarity to the secondary forest reference three years into the future (delta-t = +3) and (2) predict the restoration strategy (Natural Generation with Management, Natural Generation without Management, Full-Area Planting, Agroforestry Systems or Not Identified). We use 5-fold spatial cross-validation with k-means clustering on coordinates to ensure the folds are geographically separated. We compare Logistic Regression and Random Forest (100 trees) across several feature sets (see Figure \ref{fig:prediction_tasks}). As expected, we observe that embeddings significantly improve the accuracy of predictions about future similarity. However, using embeddings to predict the restoration strategy does not improve performance compared to using environmental and spectral signals alone. Notably, there is little difference in performance when using AlphaEarth embeddings versus NDVI/EVI alone, suggesting that the embeddings already incorporate these spectral indices indirectly.

\section{Discussion}

We propose a novel framework for using Earth observation foundation models to characterize forest restoration projects in the Brazilian Atlantic Forest. We demonstrate that AlphaEarth embeddings can form meaningful clusters based on LULC. Our ``Reference Trajectory Embedding'' approach uses cosine similarities in the embedding space between ongoing restoration projects and established secondary forests to quantify and track restoration progress.

However, AlphaEarth Foundations is a proprietary model and it remains to be seen whether the observed patterns can be generalized to open foundation models such as Clay or SatMAE. Benchmarking across architectures and biomes is an important area for future research. Additionally, the short time window (2017–2024) only captures the initial stages of restoration. Furthermore, it remains unclear whether clustering in embedding space can improve the metadata of projects with missing values. Finally, although remote-sensing methods can be useful, they should be regarded as complementary to, rather than a replacement for, community engagement.

\section*{Acknowledgments}
We thank the Observatorio da Restauração e Reflorestamento (ORR) for providing restoration polygon data. Thank you also to Hilary Brumberg and Professor Ahmed Ragab for giving valuable thoughts and insights.

\textbf{Use of LLMs}: We used Claude Code to help develop the code for Google Earth Engine scripts, experimental scripts, and data exploration and figures notebooks. We also used Google Gemini to help format tables and references in LaTeX. We used Google Gemini and Grammarly to identify spelling mistakes and improve the writing in this paper.

\newpage
\bibliography{iclr2026_conference}
\bibliographystyle{iclr2026_conference}

\newpage
\appendix

\section{Appendix}
\subsection{Data Sources}\label{sec:appendix_a}
\begin{table}[H]
\small
\centering
\caption{Data sources and derived variables used in this study.}
\label{tab:data_sources}
\begin{tabular}{@{}p{2.2cm}p{4.8cm}p{5.5cm}@{}}
\toprule
\textbf{Variable} & \textbf{Data Source} & \textbf{Details} \\
\midrule
NDVI & COPERNICUS/S2\_SR\_HARMONIZED & 2017-2024; cloud-masked. \\
\addlinespace
LULC Coverage & MapBiomas Brazil Collection 10 & Coverage V1 and Deforestation Secondary Vegetation V2; 30\,m resolution.\\
\addlinespace
Topography & USGS/SRTMGL1\_003 & Elevation, slope, aspect; 30\,m resolution. \\
\addlinespace
Climate & IDAHO\_EPSCOR/TERRACLIMATE & Annual precip, temp (max/min), evapotranspiration. \\
\addlinespace
AlphaEarth Embs & GOOGLE/SATELLITE\_EMBEDDING/V1 & 64-dim annual composites (2017-2024). \\
\bottomrule
\end{tabular}
\end{table}

\subsection{Polygon Metadata}\label{sec:appendix_b}
\begin{figure}[H]
\centering
\includegraphics[width=\textwidth]{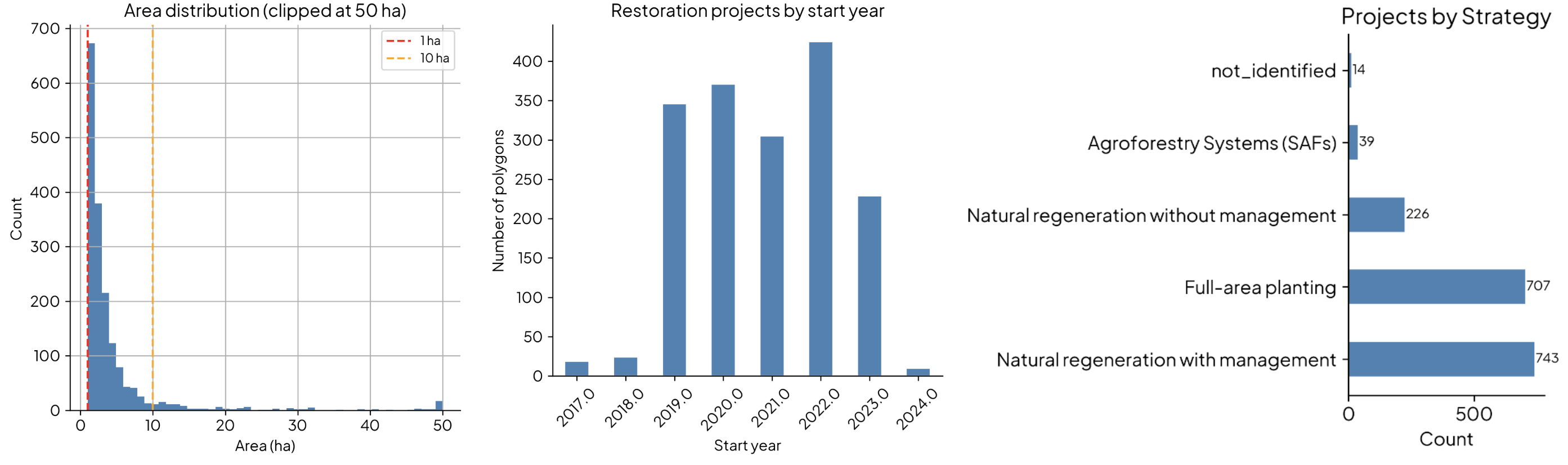}
\caption{Restoration sites metadata distributions.}
\label{fig:polygondata}
\end{figure}

Figure \ref{fig:polygondata} shows the distribution of key restoration site metadata.

\subsection{Additional Trajectory Curves}\label{sec:appendix_c}
\begin{figure}[H]
\centering
\includegraphics[width=\textwidth]{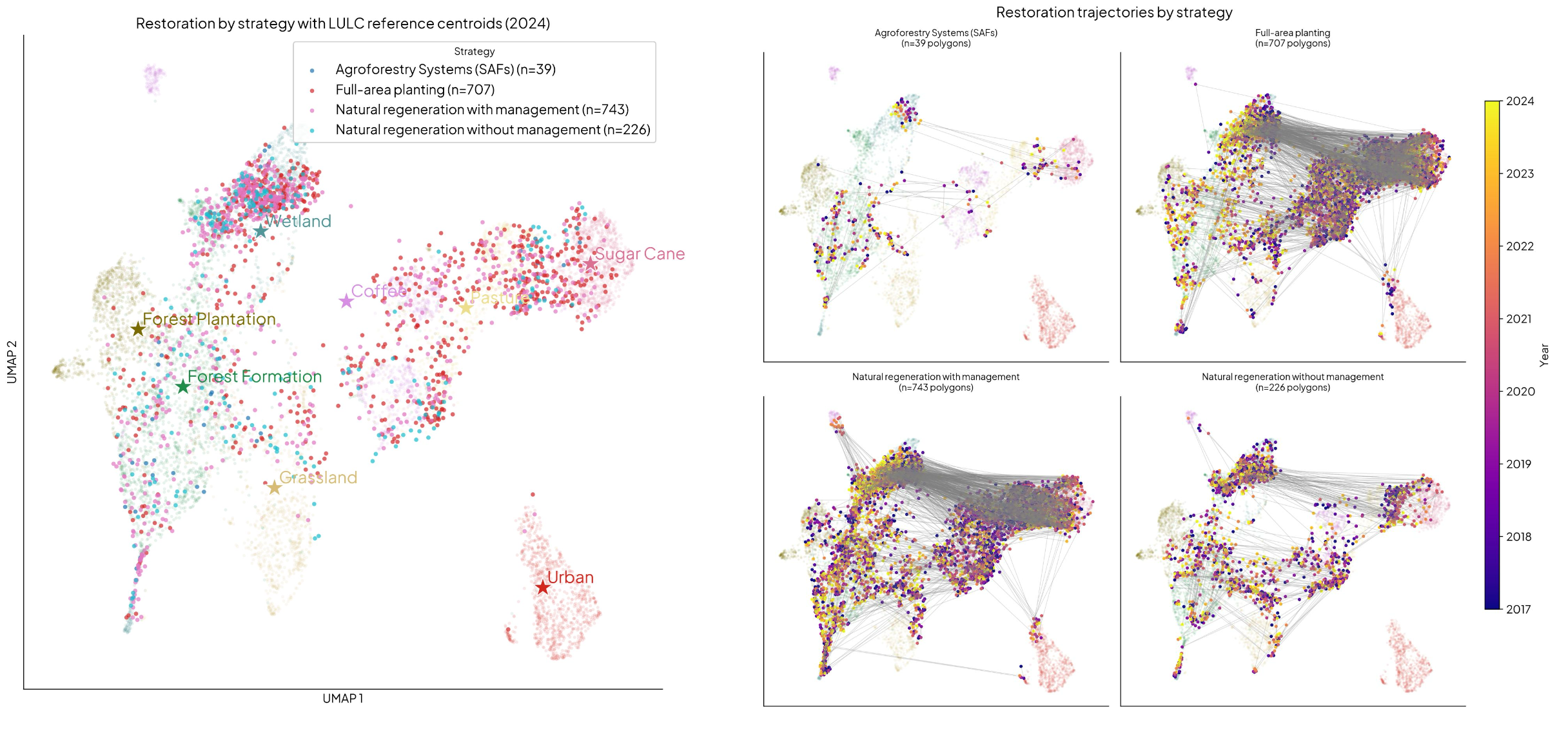}
\caption{Polygon trajectories in UMAP space by restoration strategy.}
\label{fig:umapstrategy}
\end{figure}

\begin{figure}[H]
\centering
\includegraphics[width=\textwidth]{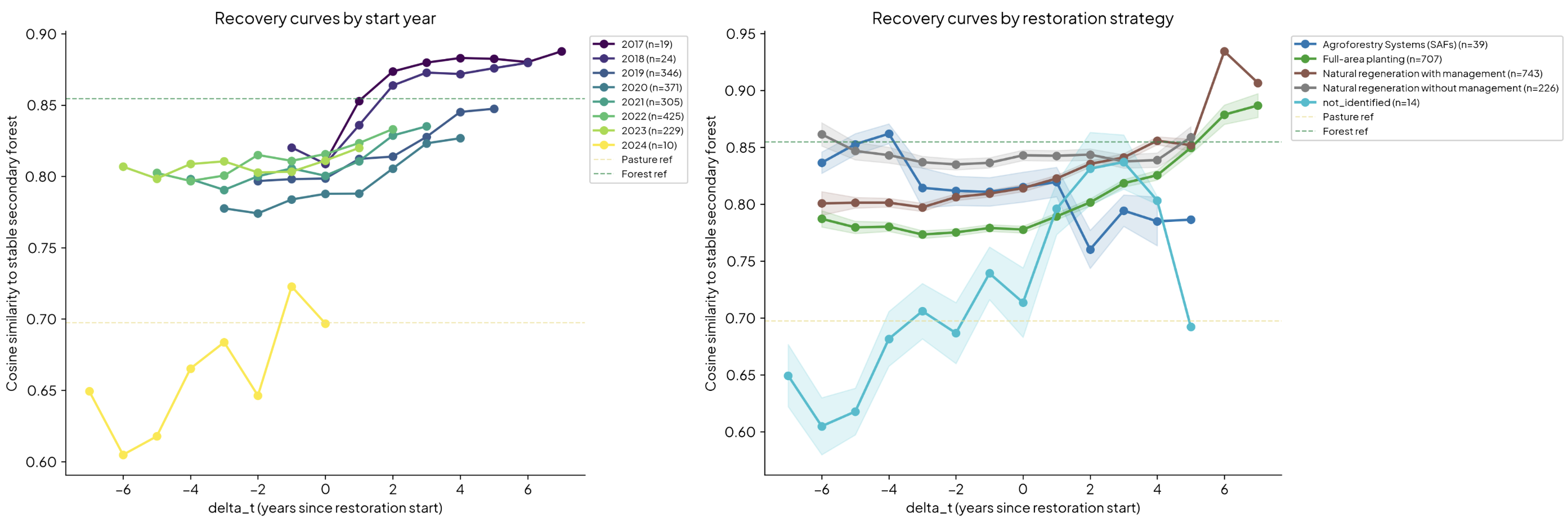}
\caption{Recovery trajectory curves stratified by start year and strategy.}
\label{fig:recoverystartstrategy}
\end{figure}

Figure \ref{fig:umapstrategy} shows the trajectories of polygons in UMAP embedding space, categorized by restoration strategy. Figure \ref{fig:recoverystartstrategy} shows recovery curves, stratified by start year and strategy.

\subsection{Trajectory Case Studies}\label{sec:appendix_d}
\begin{figure}[H]
\centering
\includegraphics[width=\textwidth]{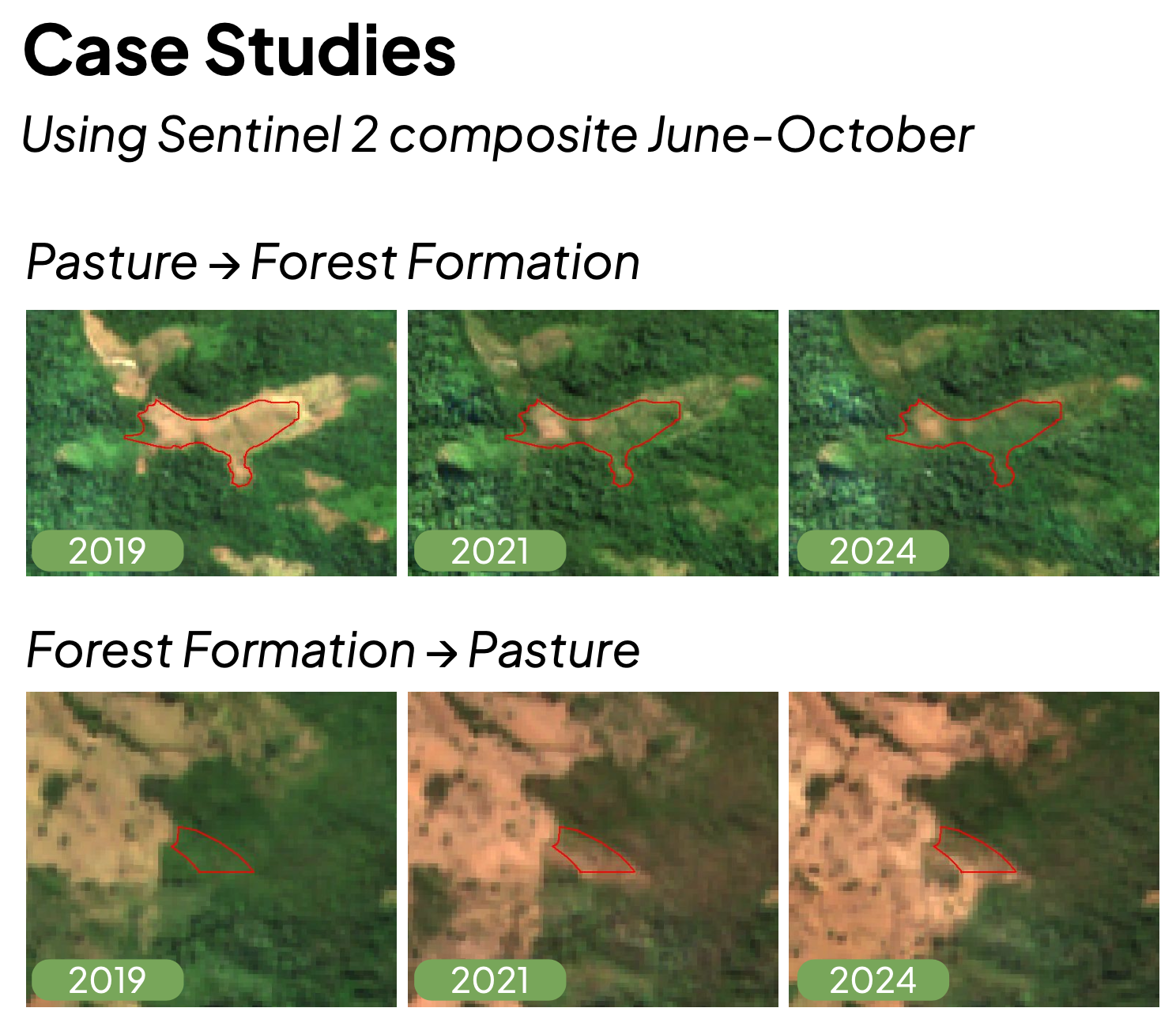}
\caption{Two case studies of identified restoration polygon transitions using embedding trajectories.}
\label{fig:case-study}
\end{figure}

Figure \ref{fig:case-study} shows two case studies of restoration polygons with identified LULC transitions using embedding trajectories.

\subsection{LULC Outliers}\label{sec:appendix_e}
\begin{figure}[H]
\centering
\includegraphics[width=\textwidth]{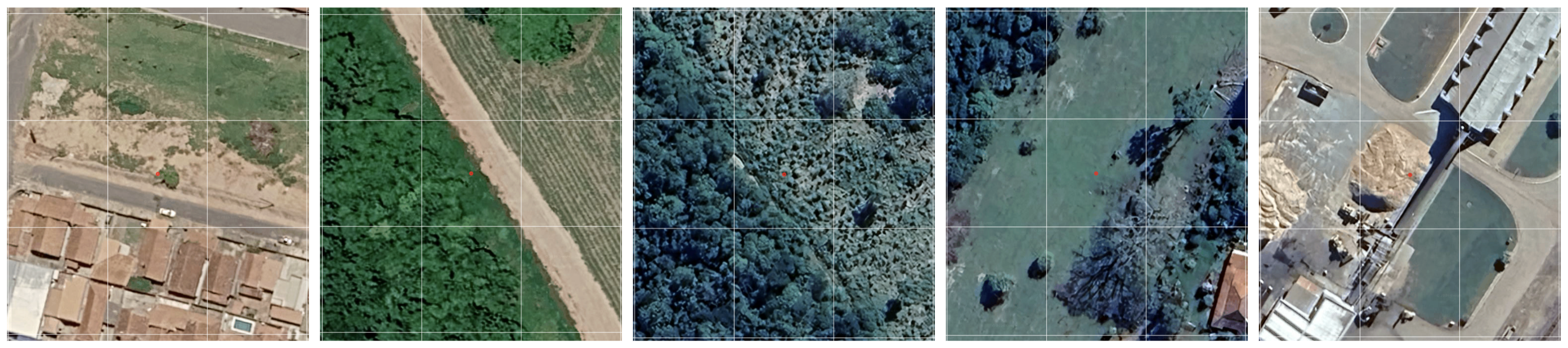}
\caption{The five points classified as 'Forest Formation' in the MapBiomas Collection 10 Coverage V1 layer, whose embeddings were the farthest from the 'Forest Formation' centroid embedding.}
\label{fig:misclass}
\end{figure}

We create LULC cluster centroids as the mean AlphaEarth embedding of all embedded points for a given stable LULC class. We then identify the points at the largest distances from their proposed centroid to detect potential misclassifications (see the case study in Figure \ref{fig:misclass}). Therefore, embeddings could be used to identify and correct potential LULC misclassifications, thereby increasing the accuracy of a LULC classification layer.

\end{document}